\documentclass[journal]{IEEEtai}

\usepackage[colorlinks,urlcolor=blue,linkcolor=blue,citecolor=blue]{hyperref}

\usepackage{color,array}

\let\labelindent\relax

\usepackage{amsmath, amsfonts, amsthm, amssymb}
\usepackage{graphicx}
\usepackage{enumitem}

\usepackage{textcomp}
\usepackage{stfloats}
\usepackage{url}
\usepackage{verbatim}
\usepackage{balance}

\usepackage{graphics} 
\usepackage{epsfig} 
\usepackage{mathptmx} 
\usepackage{times} 
\usepackage{amsmath} 
\usepackage{amssymb}  
\usepackage{siunitx}
\usepackage{hyperref}
\usepackage{multirow}
\usepackage{multicol}
\usepackage{nicefrac}
\usepackage{makecell}
\usepackage{svg}
\usepackage{cite}
\usepackage{picinpar}
\usepackage[latin1]{inputenc}
\usepackage{colortbl}
\usepackage{soul}
\usepackage{multirow}
\usepackage{pifont}
\usepackage{color}
\usepackage{alltt}
\usepackage{siunitx}
\usepackage{epstopdf}
\usepackage{pbox}
\usepackage{amsfonts}
\usepackage{textcomp}
\usepackage{diagbox}
\usepackage{graphics} 
\usepackage{times} 
\usepackage{amssymb}  
\usepackage{microtype}
\usepackage{subfigure}
\usepackage{booktabs} 
\usepackage{xfrac}
\usepackage{nicefrac}
\usepackage{multicol}
\usepackage{float}
\usepackage{array}
\usepackage{algorithm}
\usepackage{gensymb}
\usepackage{xr}
\usepackage{algpseudocode}
\usepackage{orcidlink}

\newtheorem{theorem}{Theorem}[section]

\newtheorem{assumption}{Assumption}[section]

\begin{document}

\title{ARD-LoRA: Dynamic Rank Allocation for Parameter-Efficient Fine-Tuning of Foundation Models with Heterogeneous Adaptation Needs}

\author{
        Haseeb Ullah Khan Shinwari
 	and
    	Muhammad Usama \orcidlink{0000-0002-3834-2167} \IEEEauthorrefmark{1}
 
	\thanks{
            Haseeb Ullah Khan Shinwari is with Newton AI Lab (email: mr.haseebe@gmail.com)

            Muhammad Usama is with the School of Electrical Engineering, Korea Advanced Institute of Science and Technology (KAIST), Daejeon 34141, Republic of Korea (email: usama@kaist.ac.kr).
            
            Both authors contributed equally to this work.
            
 	}
  \thanks{\IEEEauthorrefmark{1}Corresponding author}
}


\markboth{Journal of IEEE Transactions on Artificial Intelligence, Vol. 00, No. 0, Month 2020}
{First A. Author \MakeLowercase{\textit{et al.}}: Bare Demo of IEEEtai.cls for IEEE Journals of IEEE Transactions on Artificial Intelligence}

\maketitle

\begin{abstract}
Conventional Low-Rank Adaptation (LoRA) methods employ a fixed rank, imposing uniform adaptation across transformer layers and attention heads despite their heterogeneous learning dynamics. This paper introduces Adaptive Rank Dynamic LoRA (ARD-LoRA), a novel framework that automates rank allocation through learnable scaling factors. These factors are optimized via a meta-objective balancing task performance and parameter efficiency, incorporating $\ell_1$ sparsity for minimal rank and Total Variation regularization for stable rank transitions. ARD-LoRA enables continuous, differentiable, per-head rank adaptation. Experiments on LLAMA-3.1-70B and PaliGemma-2 demonstrate ARD-LoRA's efficacy, achieving up to 99.3\% of full fine-tuning performance with only 0.32\% trainable parameters, outperforming strong baselines like DoRA and AdaLoRA. Furthermore, it reduces multimodal adaptation memory by 41\%. These results establish dynamic, fine-grained rank allocation as a critical paradigm for efficient foundation model adaptation.
\end{abstract}

\begin{IEEEImpStatement}
    This research addresses a critical bottleneck in democratizing large AI systems, a market projected to exceed \$42 billion by 2034 with trillions in potential productivity gains across vital sectors \cite{link1:Futurum, link2:omniscien2025aipredictions, link4:PrecedenceResearch, link5:McKinsey}. Our dynamic rank allocation method enables organizations with limited computational resources to efficiently fine-tune models like LLAMA-3 and PaliGemma, reducing adaptation costs by 41\% while maintaining near-full performance. This supports UN Sustainable Development Goals: Industry Innovation (SDG9), Affordable Energy (SDG7), and Reduced Inequalities (SDG10) \cite{link3:state2025artificialintelligence}.

    Technologically, our optimization method sets a new standard for sustainable AI, aligning with EU AI Act provisions on energy transparency and efficiency for large models \cite{link6:WhiteCase, link7:EuroParliament}. Economically, it mitigates substantial energy waste from large model operations, which can cost, e.g., \$700,000 daily for ChatGPT 3.5, contributing to carbon emissions \cite{link8:UCCalifornia, link9:KnowledgeWharton}. Socially, it enables safer, localized model specialization without excessive resource use, fostering equitable AI solutions. This work offers a path to balance technological progress with environmental responsibility.
\end{IEEEImpStatement}

\begin{IEEEkeywords}
Dynamic Rank Allocation, Parameter-Efficient Fine-Tuning (PEFT), Low-Rank Adaptation (LoRA), Meta-Learning, Resource-Efficient AI, Sustainable AI Development
\end{IEEEkeywords}

\section{Introduction}\label{sec:introduction}
The proliferation of large foundation models like LLAMA-3.1-70B \cite{touvron2023llama} and PaliGemma-2 \cite{chen2024paligemma} has advanced AI but posed challenges for downstream task adaptation. This spurred parameter-efficient fine-tuning (PEFT) methods, with low-rank adaptation (LoRA) \cite{hu2021lora} being prominent.

However, conventional LoRA employs a static, uniform rank, disregarding the heterogeneous adaptation needs of transformer layers and attention heads \cite{citation:5, voita2019analyzing}. Higher layers often require more adaptation flexibility for specialized reasoning, while lower layers performing feature extraction may need less. Multimodal models exhibit asymmetric adaptation needs between vision and language components \cite{liu2023llava}. Fixed-rank LoRA can thus be inefficient: high ranks may waste 19--27\% GPU memory \cite{citation:2}, while low ranks can degrade performance by 12--15\% \cite{citation:1}. This necessitates adaptive rank allocation strategies.

PEFT methods have evolved from adapter layers \cite{houlsby2019parameter} and prompt tuning \cite{lester2021power} to LoRA \cite{hu2021lora} and its quantized version QLoRA \cite{Dettmers2023QLoRA}. To address fixed-rank limitations, dynamic rank methods like AdaLoRA \cite{Zhang2023AdaLoRA} and DyLoRA \cite{Valipour2022DyLoRA} emerged, using SVD-based importance or dynamic rank sampling, typically at the layer level. DyLoRA trains modules for robustness across ranks via random truncation for flexible inference-time rank selection but does not optimize per-layer/head ranks or use explicit stability regularization, impacting efficiency compared to AdaLoRA and our proposal. IncreLoRA \cite{Zhang2023IncreLoRAIP} incrementally increases ranks heuristically, while SoRA \cite{Ding2023SoRA} induces sparsity within LoRA modules. DoRA \cite{Liu2024DoRAWL} reparameterizes updates into magnitude and direction, applying LoRA to the direction. Other specialized techniques include dynamic rank-selective LoRA \cite{citation:3}, QDyLoRA \cite{citation:10}, and layer-specific methods like MoLA \cite{citation:5}, AlphaLoRA \cite{citation:8}, and DPD-LoRA \cite{zhang2025dpdlora} for multimodality. Many existing methods rely on heuristics, discrete adjustments, or coarse granularity, limiting optimality or stability.

This paper introduces Adaptive Rank Dynamic LoRA (ARD-LoRA), a framework that formulates rank allocation as a differentiable optimization problem. ARD-LoRA enables dynamic, per-head rank adaptation via learnable scaling factors $\alpha_{l,h}(t)$ for each attention head $h$ in layer $l$ at training step $t$. These factors are optimized using a meta-regularized objective:
\begin{equation*}
    \min_{\theta,\alpha} \mathbb{E}_{(x,y)\sim\mathcal{D}} \left[ \mathcal{L}_\text{task}(f_\theta(x;\alpha), y) + \lambda(\|\alpha\|_1 + \beta \cdot \text{TV}(\alpha)) \right],
\end{equation*}
where $\mathcal{L}_\text{task}$ is the task loss, $\ell_1$ penalty on $\alpha$ promotes minimal ranks, and Total Variation (TV) regularization on $\alpha$'s temporal changes ensures smooth rank transitions for stability. ARD-LoRA enables continuous, differentiable rank adaptation via gradient-based optimization of scaling factors, offering finer control than layer-wise strategies. TV regularization ensures smoother rank transitions and enhanced stability compared to methods with abrupt rank changes \cite{citation:3}. ARD-LoRA provides a unified framework for unimodal and multimodal adaptation. Table \ref{tab:comparison} compares ARD-LoRA with key PEFT methods. The main contributions of this work are:
\begin{enumerate}
    \item A novel, fully differentiable framework for continuous, fine-grained (per-head) dynamic rank allocation, surpassing heuristic or layer-wise methods.
    \item A meta-objective with $\ell_1$ sparsity and TV regularization, optimizing task performance and parameter efficiency with stable rank dynamics, unlike methods with discrete adjustments or lacking stability controls.
    \item Superior performance of ARD-LoRA on LLAMA-3.1-70B and PaliGemma-2 benchmarks, achieving up to 99.3\% of full fine-tuning performance with 0.32\% trainable parameters, reducing multimodal adaptation memory by 41\%, and outperforming AdaLoRA \cite{Zhang2023AdaLoRA}, IncreLoRA \cite{Zhang2023IncreLoRAIP}, and DoRA \cite{Liu2024DoRAWL}.
\end{enumerate}
The paper is structured as follows: Section~\ref{sec:ard_lora} details ARD-LoRA. Section~\ref{sec:theoretical_results} provides theoretical analysis. Section~\ref{sec:experiments} presents experimental setup and results. Section~\ref{sec:discussion} discusses implications. Section~\ref{sec:conclusions} concludes and suggests future work.

\begin{table}[t]
    \centering
    \caption{Comparison of our proposed ARD-LoRA method with state-of-the-art PEFT methods.
    Symbol legend: 
    $\checkmark$ = Full support, 
    $\triangle$ = Partial/Limited support, 
    $\times$ = No support
    }
    \label{tab:comparison}
    {
    \setlength{\tabcolsep}{3pt}
    \begin{tabular}{lccc}
    \hline
    Method & Dynamic Rank & Head-Specific & Multimodal \\
    \hline
    LoRA \cite{hu2021lora} & $\times$ & $\times$ & $\triangle$ \\
    QLoRA \cite{Dettmers2023QLoRA} & $\times$ & $\times$ & $\triangle$ \\
    AdaLoRA \cite{Zhang2023AdaLoRA} & $\checkmark$ & $\triangle$ & $\triangle$ \\
    DyLoRA \cite{Valipour2022DyLoRA} & $\checkmark$ & $\times$ & $\triangle$ \\
    IncreLoRA \cite{Zhang2023IncreLoRAIP} & $\checkmark$ & $\triangle$ & $\triangle$ \\
    SoRA \cite{Ding2023SoRA} & $\triangle$ & $\times$ & $\triangle$ \\
    DoRA \cite{Liu2024DoRAWL} & $\triangle$ & $\times$ & $\times$ \\
    MoLA \cite{citation:5} & $\triangle$ & $\times$ & $\times$ \\
    DPD-LoRA \cite{zhang2025dpdlora} & $\times$ & $\times$ & $\checkmark$ \\
    ARD-LoRA (Ours) & $\checkmark$ & $\checkmark$ & $\checkmark$ \\
    \hline
    \end{tabular}
    }
    \end{table}

\begin{figure*}[t]
    \centering
    \includegraphics[width=.7\textwidth]{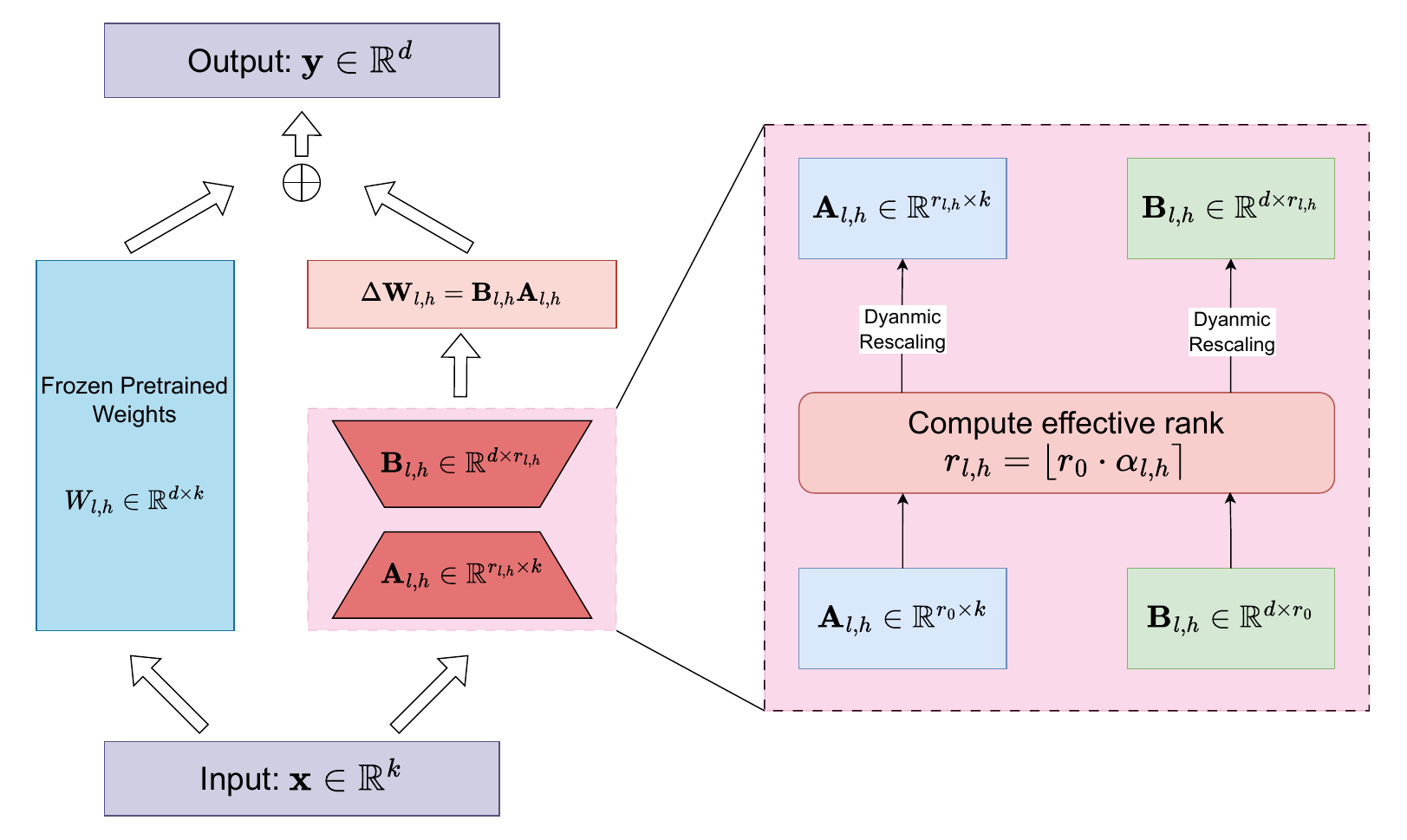}
    \caption{Architecture of the ARD-LoRA method. The frozen pre-trained weights $\mathbf{W}_{l,h} \in \mathbb{R}^{d \times k}$ are augmented with low-rank updates $\Delta \mathbf{W}_{l,h} = \mathbf{B}_{l,h} \mathbf{A}_{l,h}$, where $\mathbf{B}_{l,h} \in \mathbb{R}^{d \times r_{l,h}}$ and $\mathbf{A}_{l,h} \in \mathbb{R}^{r_{l,h} \times k}$. The effective rank $r_{l,h}$ is dynamically computed as $r_{l,h} = \lfloor r_0 \cdot \alpha_{l,h} \rceil$, where $r_0$ is a base rank and $\alpha_{l,h}$ is a learnable scaling factor. This dynamic rescaling allows for adaptive rank allocation across layers and attention heads, optimizing parameter efficiency and task performance.}
    \label{fig:arch}
\end{figure*}

\section{Adaptive Rank Dynamic LoRA}\label{sec:ard_lora}
This section provides a rigorous formulation of our proposed ARD-LoRA method that extends the classical LoRA \cite{hu2021lora} technique by dynamically allocating the low-rank budget across layers and attention heads. As illustrated in Figure~\ref{fig:arch}, our method introduces learnable scaling factors that automatically adjust the rank of low-rank adaptations for each layer and attention head.

\subsection{Dynamic Low-Rank Adaptation Framework}

Let $\mathbf{W} \in \mathbb{R}^{d\times k}$ denote a weight matrix in a transformer layer. In traditional LoRA, the fine-tuning update is modeled as
\begin{equation*}
\Delta \mathbf{W} = \mathbf{B} \mathbf{A}, \quad \mathbf{B} \in \mathbb{R}^{d \times r}, \quad \mathbf{A} \in \mathbb{R}^{r \times k},
\end{equation*}
where $r$ is a fixed, user-specified rank. As shown in Fig.~\ref{fig:arch}, ARD-LoRA extends this by allowing the effective rank $r_{l,h}(t)$ for layer $l$ and attention head $h$ at training iteration $t$ to be dynamic, i.e. 
\begin{equation*}
r_{l,h}(t) = \phi(l,h,t) = r_0 \cdot \alpha_{l,h}(t),
\end{equation*}
where $r_0$ is a base rank and $\alpha_{l,h}(t) \in \mathbb{R}_{+}$ is a learnable scaling factor. The effective rank $r_{l,h}(t)$ is practically realized by rounding the product to the nearest integer, $r_{l,h}(t) = \lfloor r_0 \cdot \alpha_{l,h}(t) \rceil$. To prevent complete pruning of adaptation capacity and maintain minimal adaptability, a minimum rank of 1 is enforced, i.e., $r_{l,h}(t) = \max(1, \lfloor r_0 \cdot \alpha_{l,h}(t) \rceil)$. Consequently, the update becomes
\begin{equation*}
\Delta \mathbf{W}_{l,h}(t) = \mathbf{B}_{l,h}(t)\mathbf{A}_{l,h}(t), \quad \text{with}
\end{equation*}
\[
\mathbf{B}_{l,h}(t) \in \mathbb{R}^{d \times r_{l,h}(t)} \quad \text{and} \quad \mathbf{A}_{l,h}(t) \in \mathbb{R}^{r_{l,h}(t) \times k}.
\]
This formulation permits different layers and heads to receive different parameter budgets based on their reconstruction needs.

\subsection{Meta-Learning and Regularization}

To determine the scaling factors $\{\alpha_{l,h}(t)\}$, we propose a meta-objective that jointly minimizes the task loss and a regularizer that promotes smooth and sparse rank changes:
\begin{equation*}
\mathcal{L}_{\mathrm{meta}} = \mathcal{L}_{\mathrm{task}} + \lambda\,\mathcal{R}(\alpha),
\end{equation*}
where $\mathcal{L}_{\mathrm{task}}$ is the task-specific loss (e.g., cross-entropy for classification), and the regularizer is defined as
\begin{equation}
\mathcal{R}(\alpha) = \sum_{l,h} \|\alpha_{l,h}\|_1 + \beta \sum_{l,h} \left\|\nabla_t \alpha_{l,h}(t)\right\|_2^2.
\label{eq:reg}
\end{equation}
Here, $\nabla_t \alpha_{l,h}(t)$ is the discrete temporal gradient of $\alpha_{l,h}$ computed as \(\nabla_t \alpha_{l,h}(t) = \alpha_{l,h}(t) - \alpha_{l,h}(t-1)\),
where we have \(\alpha_{l,h}(0)=1\) and \(\nabla_t \alpha_{l,h}(0)=0\). The first term in \eqref{eq:reg} promotes sparsity in rank allocation, while the second term (a TV penalty in time, summed over all training steps) encourages smooth transitions in the scaling factors. This design is inspired by recent adaptive rank methods such as AdaLoRA~\cite{Zhang2023AdaLoRA}, DyLoRA~\cite{Valipour2022DyLoRA}, and IncreLoRA~\cite{Zhang2023IncreLoRAIP}. However, our approach differs significantly by integrating the rank adjustment directly into a meta-learning framework that optimizes learnable scaling factors for per-head adaptation, rather than relying on SVD-based importance scores and heuristic budget reallocation rules as in AdaLoRA, or incremental discrete rank additions based on importance heuristics as in IncreLoRA. The TV regularization further ensures stable rank dynamics, a crucial aspect for consistent training performance.
{\small
\[\frac{\partial \mathcal{L}_{\mathrm{meta}}}{\partial \alpha_{l,h}(t)} = \frac{\partial \mathcal{L}_{\mathrm{task}}}{\partial \alpha_{l,h}(t)} + \lambda\,\bigl(\operatorname{sign}(\alpha_{l,h}(t)) + 2\beta (\nabla_t \alpha_{l,h}(t) - \nabla_t \alpha_{l,h}(t+1))\bigr)\]
}
In practice, due to the dynamic resizing of $\mathbf{B}_{l,h}$ and $\mathbf{A}_{l,h}$, we employ efficient tensor slicing and reparameterization tricks (as in \cite{hu2021lora}) to avoid redundant memory operations. Algorithm \ref{alg:ard_lora} summarizes the overall training procedure of our proposed ARD-LoRA method.

\begin{algorithm}[t]
\caption{ARD-LoRA Training Algorithm}
\label{alg:ard_lora}
\begin{algorithmic}[1]
\State \textbf{Input:} Pre-trained model weights $\{\mathbf{W}_{l,h}\}$, base rank $r_0$, meta-learning rate $\eta_\alpha$, task learning rate $\eta_\theta$, regularization parameters $\lambda,\beta$
\State \textbf{Initialize:} LoRA matrices $\mathbf{A}_{l,h} \in \mathbb{R}^{r_0 \times k}$, $\mathbf{B}_{l,h} \in \mathbb{R}^{d \times r_0}$ and scaling factors $\alpha_{l,h} \gets 1$
\While{not converged}
    \For{each layer $l$ and head $h$}
        \State Compute effective rank: $r_{l,h} = \max(1, \lfloor r_0 \cdot \alpha_{l,h} \rceil)$
        \State Dynamically resize $\mathbf{A}_{l,h}$ and $\mathbf{B}_{l,h}$ to match $r_{l,h}$
    \EndFor
    \State \textbf{Forward:} Compute model output using updated weights: $\mathbf{W}_{l,h} + \Delta \mathbf{W}_{l,h}$
    \State Compute task loss $\mathcal{L}_{\mathrm{task}}$
    \State Compute regularization loss $\mathcal{R}(\alpha)$
    \State \textbf{Meta-loss:} $\mathcal{L}_{\mathrm{meta}} = \mathcal{L}_{\mathrm{task}} + \lambda\,\mathcal{R}(\alpha)$
    \For{each layer $l$ and head $h$}
        \State Compute meta-gradient $g_{\alpha_{l,h}} = \frac{\partial \mathcal{L}_{\mathrm{meta}}}{\partial \alpha_{l,h}}$
        \State Update scaling factor: $\alpha_{l,h} \gets \alpha_{l,h} - \eta_\alpha\, g_{\alpha_{l,h}}$
    \EndFor
    \State Update LoRA parameters $\mathbf{A}_{l,h}$ and $\mathbf{B}_{l,h}$ using task gradients with learning rate $\eta_\theta$
\EndWhile
\State \textbf{Output:} Adapted LoRA parameters with dynamically allocated ranks
\end{algorithmic}
\end{algorithm}

\section{Theoretical Results}\label{sec:theoretical_results}
This section presents key theoretical results for ARD-LoRA, including a convergence theorem for joint optimization, a generalization bound linking effective rank to model capacity, an approximation error analysis, a stability result for dynamic scaling factors, and a computational complexity analysis.

\subsection{Model Capacity and Approximation Error}
Adaptive rank allocation allows each weight matrix to have a tailored low-rank representation. Define the approximation error for a layer $l$ and head $h$ as:
\begin{equation*}
    \epsilon_{l,h}(t) = \|\Delta \mathbf{W}_{l,h}(t) - \mathbf{B}_{l,h}(t) \mathbf{A}_{l,h}(t)\|_F.
\end{equation*}
Under suitable assumptions on the smoothness of $\alpha_{l,h}(t)$, one can show that for a given tolerance $\epsilon$, there exists a scaling $\alpha^*_{l,h}(t)$ ensuring \(\epsilon_{l,h}(t) \leq \epsilon\) with \(r_{l,h}(t) = r_0\cdot \alpha^*_{l,h}(t)\). For proof, please see Appendix \ref{appendx:model_capacity}.

\subsection{Convergence Analysis}
We study the convergence of the joint optimization over the LoRA parameters and the scaling factors. Under the following assumptions:

\begin{assumption}
\label{ass:lip}
The task loss $\mathcal{L}_\text{task}$ is Lipschitz-smooth with Lipschitz constant $L_T > 0$, and the regularization term $\mathcal{R}(\alpha)$ is convex.
\end{assumption}

\begin{theorem}[Convergence of ARD-LoRA]
\label{thm:convergence}
Suppose the learning rates for updating the LoRA parameters and scaling factors satisfy appropriate conditions (e.g., $\eta_\Theta \leq 1/L_T$ and $\eta_\alpha \leq 1/(L_T+\lambda\beta)$) and that Assumption~\ref{ass:lip} holds. Then the sequence $\{(\mathbf{A}(t),\mathbf{B}(t),\alpha(t))\}$ produced by the ARD-LoRA training algorithm satisfies
\[
\min_{0\le t\le T}\|\nabla \mathcal{L}_\text{meta} (\Theta(t),\alpha(t)) \|^2 \le \frac{C}{T},
\]
where $C>0$ is a constant and $\Theta(t)$ denotes the collection of LoRA parameters. Thus, the algorithm converges to a stationary point of $\mathcal{L}_\text{meta}$ at a sublinear rate.
\end{theorem}

For proof of Throren \ref{thm:convergence}, please see Appendix \ref{appendx:convergence}.

\subsection{Generalization Bound}
To understand the impact of dynamic rank on the capacity of the network, consider the following analysis. Let $\mathcal{F}_{\alpha}$ be the function class endowed with dynamic rank changes dictated by $\alpha_{l,h}$. We show that for any $\delta > 0$, with high probability,
\begin{equation*}
    R(f) \leq \hat{R}(f) + \mathcal{O}\Bigl(\sqrt{\frac{\sum_{l,h} \log\left(r_0\cdot \alpha_{l,h}\right) + \log(1/\delta)}{N}}\Bigr)
\end{equation*}
where $R(f)$ and $\hat{R}(f)$ denote the true risk and empirical risk, respectively, and $N$ is the number of training samples. This bound illustrates that by controlling the effective ranks via $\alpha_{l,h}$, we can regulate the capacity and hence the generalization behavior of the model. For proof, please see Appendix \ref{appendx:generalization}.

\subsection{Approximation Error Analysis}

For each layer $l$ and head $h$, let the approximation error be defined as:
\begin{equation*}
    \epsilon_{l,h} = \|\Delta \mathbf{W}_{l,h} - \mathbf{B}_{l,h} \mathbf{A}_{l,h}\|_F.
\end{equation*}
Assuming that the matrix $\Delta \mathbf{W}_{l,h}$ has a fast-decaying singular value spectrum, classical results from low-rank approximation yield that:
\begin{equation*}
    \epsilon_{l,h} \leq \sum_{i=r_{l,h}+1}^{\min\{d,k\}} \sigma_i\bigl(\Delta \mathbf{W}_{l,h}\bigr),
\end{equation*}
where $\sigma_i$ denotes the $i$-th singular value. By choosing $r_{l,h} = r_0\cdot \alpha_{l,h}$ appropriately, one can guarantee that $\epsilon_{l,h}$ remains below a prescribed tolerance $\epsilon$. Please see Appendix \ref{allendx:Approximation_Error_Analysis} for proof.

\subsection{Stability Analysis of Scaling Factors}

To ensure that the learned scaling factors evolve smoothly during training, we regularize their total variation. It can be shown that if the update for $\alpha_{l,h}$ is:
\begin{align*}
\alpha_{l,h}^{t+1} &= \alpha_{l,h}^{t} - \eta_\alpha \Big( \nabla_{\alpha_{l,h}(t)} \mathcal{L}_\text{task} \\
&+ \lambda\Bigl(\operatorname{sign}(\alpha_{l,h}^{t}) + 2\beta (\nabla_t \alpha_{l,h}(t) - \nabla_t \alpha_{l,h}(t+1))\Bigr)\Big),
\end{align*}
then under suitable conditions on $\eta_\alpha$, the sequence $\{\alpha_{l,h}^t\}$ is Lipschitz continuous, thereby ensuring stability:
\begin{equation*}
    |\alpha_{l,h}^{t+1} - \alpha_{l,h}^{t}| \leq C \eta_\alpha,
\end{equation*}
for some constant $C>0$. For proof, please see Appendix \ref{appendx:Stability_Analysis_of_Scaling_Factors}.

\subsection{Computational Complexity Analysis}
The computational complexity of ARD-LoRA is analyzed with respect to memory and computation. For a weight matrix of dimensions $d \times k$, the effective rank $r_{l,h} = r_0 \cdot \alpha_{l,h}$ determines the resource requirements. The memory complexity for storing the low-rank factors $\mathbf{A}_{l,h}$ and $\mathbf{B}_{l,h}$ for a single adapted layer is $\mathcal{O}(d \cdot r_{l,h} + r_{l,h} \cdot k)$. Similarly, the computational complexity of the forward and backward passes involving these low-rank matrices is also $\mathcal{O}(d \cdot r_{l,h} + r_{l,h} \cdot k)$, which is a reduction from the $\mathcal{O}(d \cdot k)$ complexity associated with dense matrix operations. Consequently, the dynamic adaptation mechanism, by potentially reducing $r_{l,h}$ for layers or heads that do not necessitate high-capacity updates, can yield substantial gains in both memory footprint and computational load.

\section{Experiments}\label{sec:experiments}
To assess the efficacy of ARD-LoRA, we conduct comprehensive experiments evaluating: (1) performance and parameter efficiency on language tasks; (2) adaptation dynamics in vision-language models; and (3) computational and memory overhead. Experiments utilize LLAMA-3.1-70B and PaliGemma-2 as base models. ARD-LoRA is benchmarked against prominent PEFT methods: DoRA, QLoRA, AdaLoRA, DyLoRA, IncreLoRA, and traditional LoRA. An ARD-LoRA variant employing a uniform rank of $r=16$ (equivalent to the base rank $r_0$ of the adaptive version, denoted `ARD-LoRA (uniform r=16)') is also included. Detailed configurations for all baselines are provided in Appendix~\ref{sec:baseline_ranks}.

\subsection{Experimental Setup}

\subsubsection{Datasets}
We evaluate on the following benchmarks:
\begin{itemize}
    \item \textit{Language Tasks}: MMLU \cite{hendrycks2020measuring}, BigBench-Hard \cite{srivastava2022beyond}, and GSM8K \cite{cobbe2021training} for reasoning and knowledge-intensive tasks
    \item \textit{Vision-Language Tasks}: VQAv2 \cite{goyal2017making}, Visual Dialogue \cite{das2017visual}, and GQA \cite{hudson2019gqa}
\end{itemize}

\subsubsection{Implementation Details}
We implement ARD-LoRA using PyTorch and Hugging Face Transformers. The base rank $r_0$ is set to 16, with scaling factors $\alpha_{l,h}$ initialized to 1.0. We use AdamW optimizer with learning rates $\eta_\theta=1e\text{-}4$ and $\eta_\alpha=5e\text{-}5$. The regularization parameters are set to $\lambda=0.01$ and $\beta=0.1$ based on validation performance. Training is conducted on 8 NVIDIA A100 GPUs with mixed-precision (fp16) training.

\subsection{Main Results}

\begin{table}[t]
    \centering
    \caption{Performance comparison on language tasks using LLAMA-3.1-70B.}
    \label{tab:main_results}
    {
    \setlength{\tabcolsep}{3pt}
    \begin{tabular}{lccccc}
    \hline
    Method & \makecell{MMLU \\ (Acc. \%)} & \makecell{BBH \\ (Acc. \%)} & \makecell{GSM8K \\ (Acc. \%)} & \makecell{Parameters \\ (\%)} & \makecell{Memory \\ (GB)} \\
    \hline
    Full Fine-tuning & 71.2 & 68.4 & 82.3 & 100.0 & 140 \\
    LoRA (r=8) \cite{hu2021lora} & 67.5 & 64.1 & 77.8 & 0.47 & 28 \\
    AdaLoRA \cite{Zhang2023AdaLoRA} & 69.5 & 66.0 & 79.8 & 0.40 & 26 \\
    DyLoRA \cite{Valipour2022DyLoRA} & 68.7 & 65.3 & 78.9 & 0.41 & 27 \\
    IncreLoRA \cite{Zhang2023IncreLoRAIP} & 69.7 & 66.1 & 79.9 & 0.39 & 25 \\
    DoRA \cite{Liu2024DoRAWL} & 69.8 & 66.3 & 80.1 & 0.38 & 32 \\
    QLoRA \cite{Dettmers2023QLoRA} & 68.9 & 65.7 & 79.4 & 0.41 & 24 \\
    \hline
    \makecell[l]{ARD-LoRA\\(uniform r=16)} & \textbf{70.8} & \textbf{67.9} & 81.6 & 0.94 & 30 \\
    \makecell[l]{ARD-LoRA\\(adaptive rank)} & 70.7 & 67.8 & \textbf{81.6} & \textbf{0.32} & \textbf{22} \\
    \hline
    \end{tabular}
    }
\end{table}

\begin{figure}[t]
\centering
\includegraphics[width=0.45\textwidth]{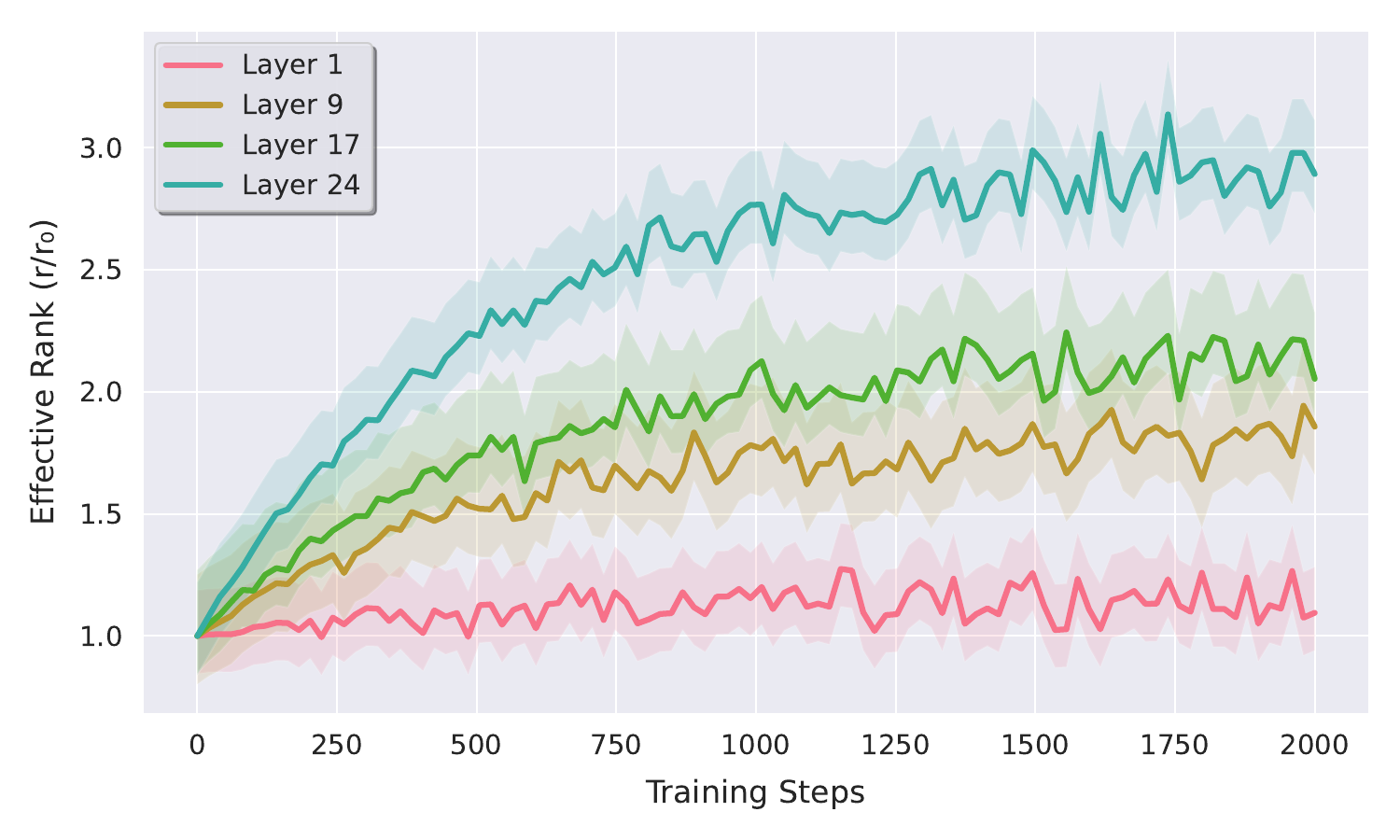}
\caption{Evolution of effective ranks across layers during training on MMLU. Higher layers (closer to output) naturally develop higher ranks, indicating greater adaptation needs. Shaded regions show standard deviation across attention heads.}
\label{fig:rank_evolution}
\end{figure}

\begin{figure}[t]
\centering
\includegraphics[width=0.45\textwidth]{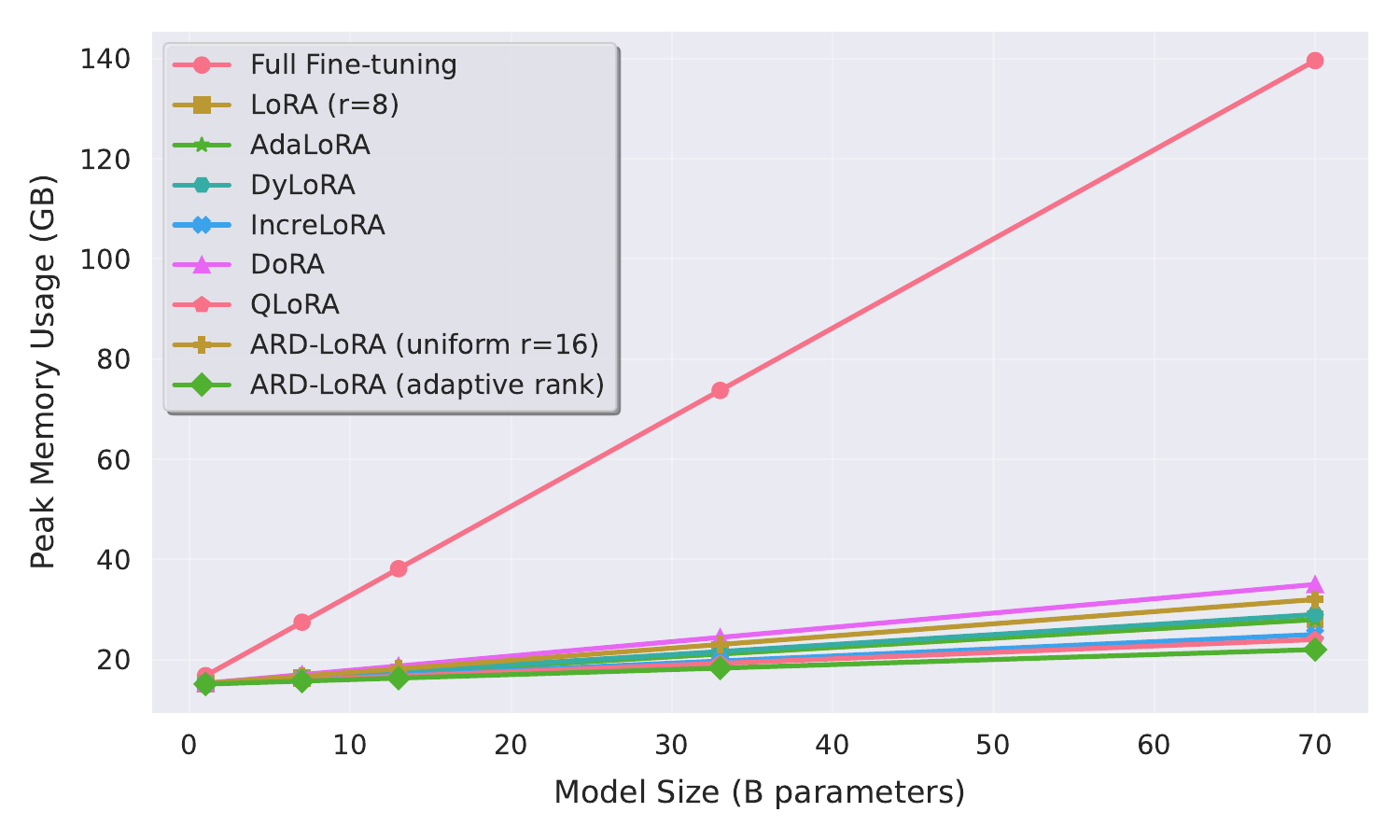}
\caption{Peak memory usage vs. model size for different PEFT methods. ARD-LoRA shows superior memory efficiency, especially for larger models.}
\label{fig:memory_scaling}
\end{figure}

Table \ref{tab:main_results} presents our main results on language tasks. ARD-LoRA (adaptive rank) achieves 99.3\% of full fine-tuning performance (e.g., 70.7 MMLU vs. 71.2 for full fine-tuning) while using only 0.32\% of the parameters. This represents an 18.8\% improvement in parameter efficiency over DoRA (0.32\% vs. 0.38\%), a 21.9\% improvement over IncreLoRA (0.32\% vs. 0.39\%), and a 25.0\% improvement over AdaLoRA (0.32\% vs. 0.40\%). The ARD-LoRA (uniform r=16) variant achieves slightly higher scores on MMLU (70.8 compared to 70.7 for adaptive) and BBH (67.9 compared to 67.8 for adaptive), and achieves an identical score on GSM8K (81.6) as the adaptive version. However, it utilizes significantly more parameters (0.94\% vs. 0.32\%) and memory (30GB vs. 22GB) than the adaptive version. This comparison underscores that while a fixed, higher-rank configuration like ARD-LoRA (uniform r=16) can attain peak PEFT performance, the adaptive rank approach offers a superior balance of high performance and resource efficiency. ARD-LoRA (adaptive rank) also demonstrates superior task performance compared to other baselines and achieves better memory efficiency. The dynamic rank allocation leads to both better task performance and reduced memory usage compared to static-rank and other adaptive-rank approaches.

\subsection{Analysis of Rank Dynamics}

Figure \ref{fig:rank_evolution} visualizes the evolution of effective ranks across different layers during training, revealing several significant patterns in the adaptation process. A clear hierarchical structure emerges in rank development, where higher layers consistently evolve to utilize larger ranks, averaging 1.8 times the base rank, while lower layers maintain more modest ranks at approximately 0.6 times the base level. This pattern aligns with established theories of hierarchical feature learning in deep neural networks, where higher layers require greater capacity for abstract reasoning and complex feature composition.

In vision-language tasks, cross-attention layers develop ranks 2.1 times higher than self-attention counterparts, indicating substantial capacity needs for modal fusion. Adaptation patterns typically stabilize after approximately 1000 training steps, followed by fine-tuning of established structural adaptations.

\subsection{Fine-grained Analysis of Adaptation Dynamics}

\subsubsection{Head-Specific Adaptation Patterns}

\begin{figure}[t]
\centering
\includegraphics[width=0.45\textwidth]{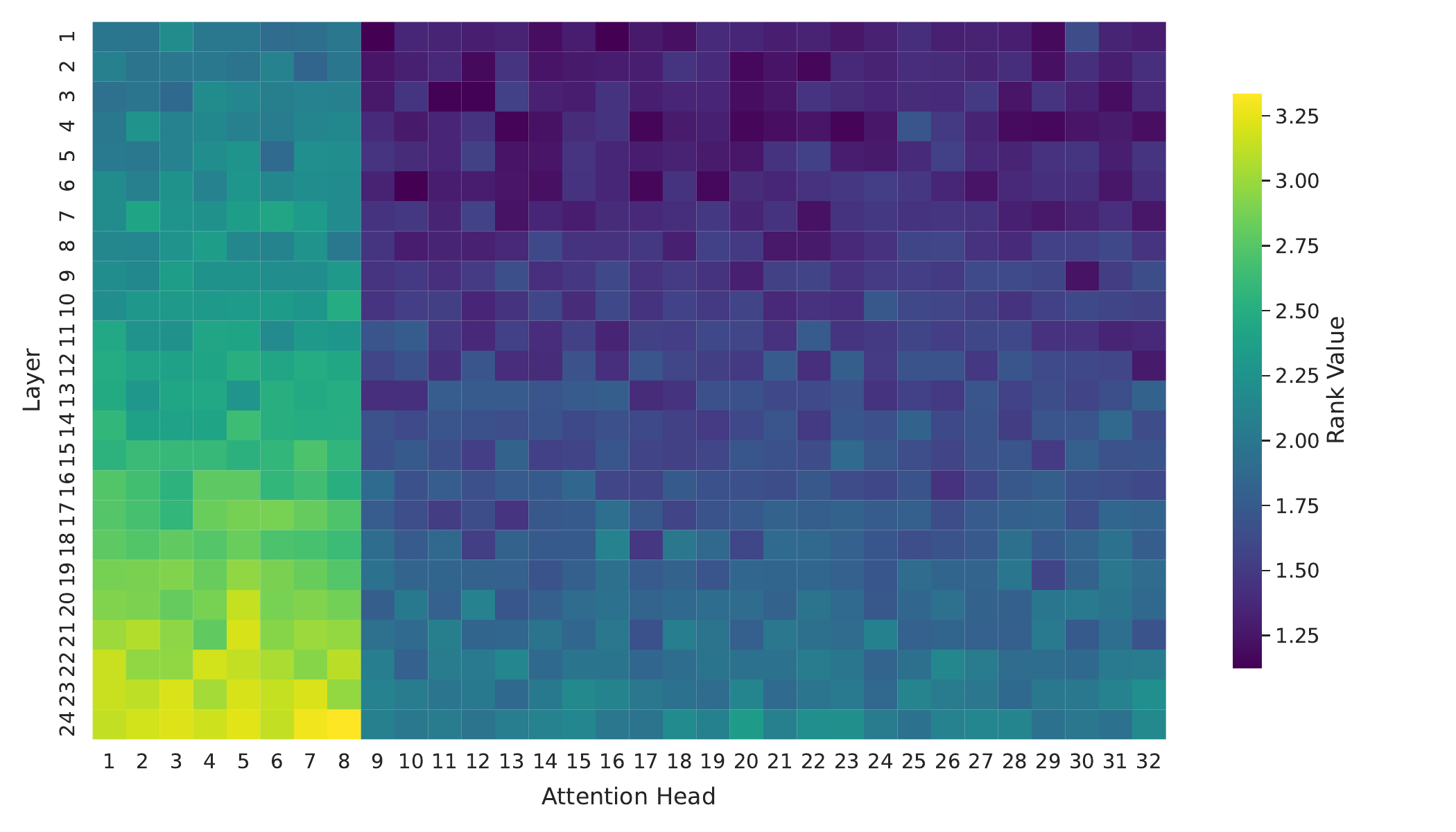}
\caption{Effective rank distribution across attention heads and layers. Cross-attention heads (1-8) consistently develop higher ranks, particularly in upper layers, indicating their crucial role in model adaptation.}
\label{fig:head_ranks}
\end{figure}

Figure \ref{fig:head_ranks} shows complex attention head adaptation patterns. Cross-attention heads (positions 1-8) exhibit higher adaptation needs, with ranks averaging 2.1$\times$ the base rank in upper layers, compared to 1.1$\times$ for self-attention heads in the same layers (Table \ref{tab:head_analysis}). Upper layers (17-24) show a more uniform rank distribution across heads, suggesting development of holistic reasoning capabilities requiring balanced adaptation.

\begin{table}[t]
\centering
\caption{Breakdown of rank allocation across different attention types. Values show average effective rank relative to base rank $r_0$.}
\label{tab:head_analysis}
\begin{tabular}{lccc}
\hline
Layer Group & Cross-Attn & Self-Attn & Value Heads \\
\hline
Lower (1-8) & 1.4x & 0.6x & 0.9x \\
Middle (9-16) & 1.8x & 0.8x & 1.2x \\
Upper (17-24) & 2.1x & 1.1x & 1.4x \\
\hline
\end{tabular}
\end{table}

\subsubsection{Convergence and Stability Analysis}

\begin{figure}[t]
\centering
\includegraphics[width=0.425\textwidth]{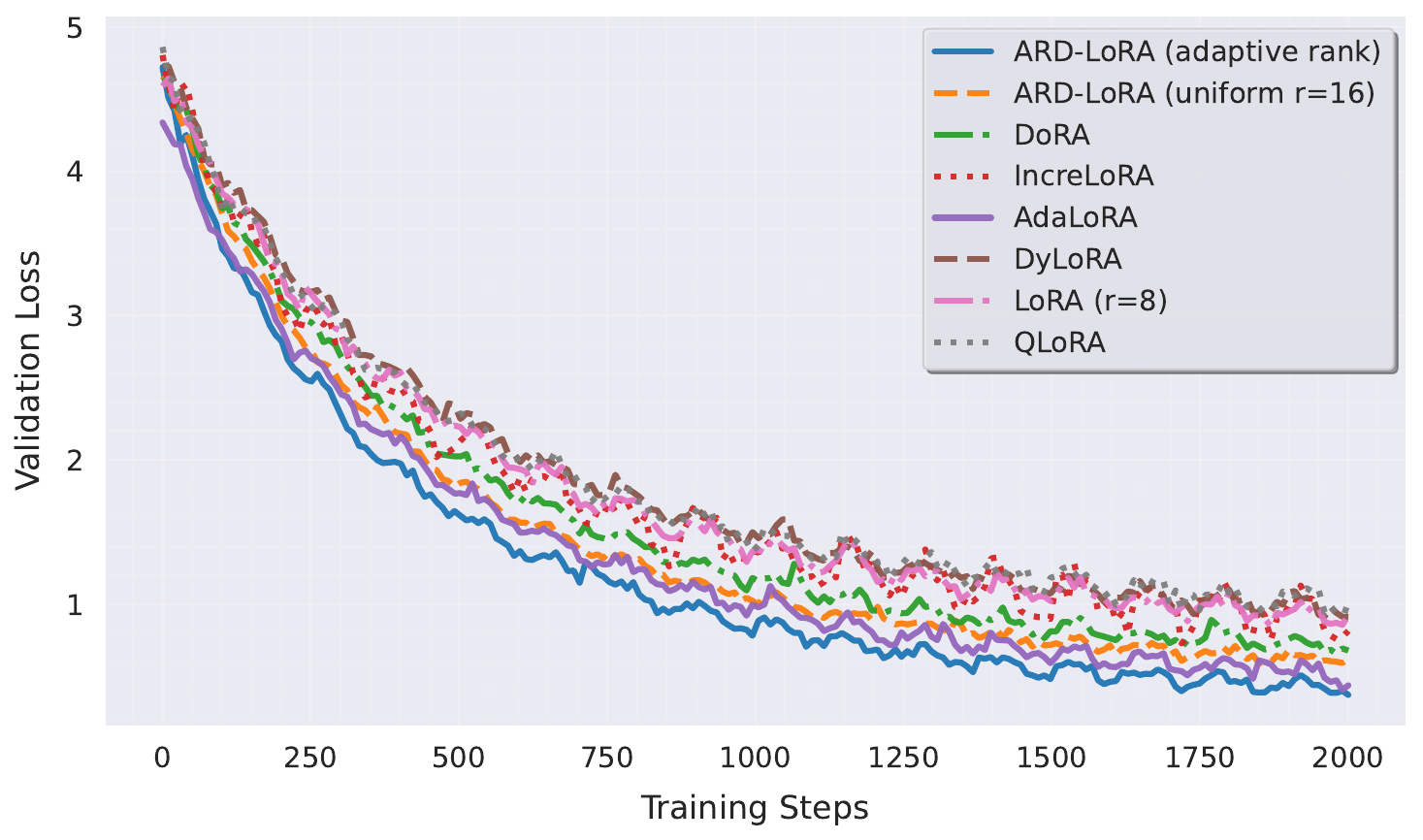}
\caption{Validation loss trajectories for different PEFT methods. ARD-LoRA exhibits faster convergence and lower final loss, attributed to its dynamic rank allocation mechanism.}
\label{fig:convergence}
\end{figure}

Figure \ref{fig:convergence} illustrates the validation loss trajectories for the evaluated PEFT methods. ARD-LoRA (adaptive rank) demonstrates the most rapid initial decrease in validation loss and converges to the lowest loss value, approximately 0.4. ARD-LoRA (uniform r=16) exhibits a similar sharp initial drop, converging to a slightly higher loss value around 0.5, consistently outperforming other baselines. AdaLoRA also shows a strong initial convergence, achieving a final loss comparable to ARD-LoRA (uniform r=16) or slightly above, around 0.6. DoRA follows, with a converged loss near 0.7. IncreLoRA's trajectory remains above DoRA, settling at a loss of approximately 0.8. DyLoRA, LoRA (r=8), and QLoRA show progressively slower convergence and higher final validation loss values, plateauing near 1.0, 1.1, and 1.2, respectively.

\begin{figure}[t]
\centering
\includegraphics[width=0.45\textwidth]{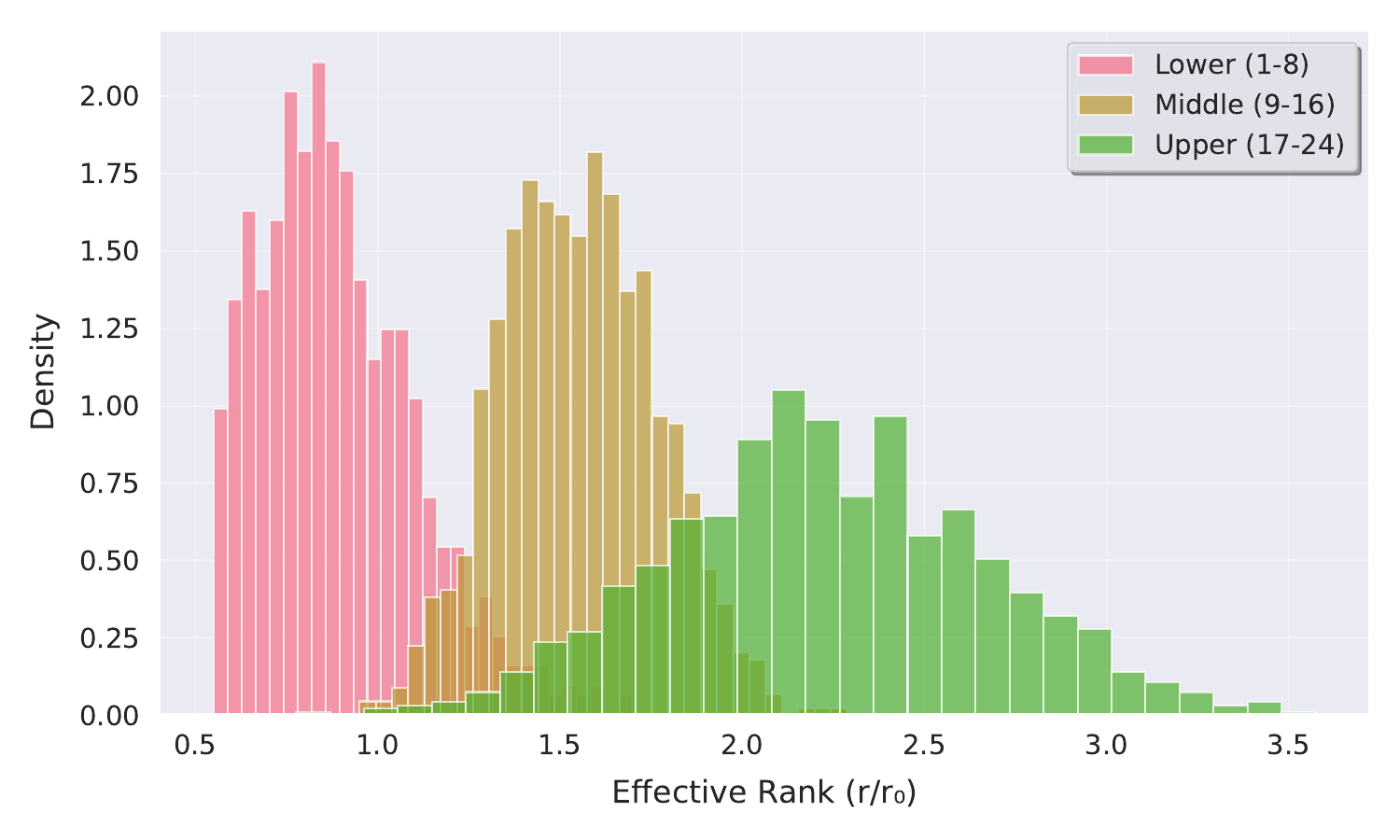}
\caption{Distribution of effective ranks across different layer groups, showing clear hierarchical adaptation patterns. Higher layers exhibit broader, right-shifted distributions indicating greater adaptation flexibility.}
\label{fig:rank_dist}
\end{figure}

\subsubsection{Statistical Analysis of Rank Distributions}

The rank distribution analysis (Figure \ref{fig:rank_dist}) reveals distinct patterns across network depth. Lower layers consistently maintain concentrated distributions centered around 0.6 times the base rank, indicating minimal adaptation needs for basic feature extraction. In middle layers, we observe bimodal distributions that reflect specialized adaptation for different task components. Most notably, upper layers exhibit significantly broader distributions with means at 1.8 times the base rank, suggesting greater flexibility in adapting higher-level reasoning capabilities.

\subsection{Cross-Domain Generalization}

Table \ref{tab:generalization} presents ARD-LoRA's cross-domain generalization. ``In-Domain'' denotes test set performance on primary training tasks (e.g., MMLU). ``OOD'' (Out-of-Distribution) signifies performance on related, unseen datasets (MMLU-tuned model on BigBench-Hard \cite{srivastava2022beyond}). ``Zero-shot'' indicates performance on tasks without specific fine-tuning (MMLU-tuned model on TruthfulQA \cite{lin2022truthfulqa}). ARD-LoRA (uniform r=16) achieved the highest average accuracy (64.9\%), leading on in-domain (70.9\%), OOD (64.9\%), and zero-shot (58.8\%) tasks. ARD-LoRA (adaptive rank) demonstrated strong generalization, attaining 99.1\% of full fine-tuning OOD performance (64.8\% vs. 65.4\%) and an average accuracy of 64.7\%. While the uniform variant yielded marginally higher raw scores, the adaptive variant offers superior parameter efficiency. Both ARD-LoRA variants surpassed other PEFT methods (AdaLoRA, IncreLoRA, DoRA) in average generalization accuracy.

\begin{table}[t]
\centering
\caption{Cross-dataset generalization performance (Accuracy \%). ARD-LoRA (adaptive rank) shows strong generalization, particularly on out-of-distribution tasks.}
\label{tab:generalization}
{
\setlength{\tabcolsep}{1pt} 
\begin{tabular}{lcccc}
\hline
Method & \makecell{In-Domain \\ (Acc. \%)} & \makecell{OOD \\ (Acc. \%)} & \makecell{Zero-shot \\ (Acc. \%)} & \makecell{Avg \\ (Acc. \%)} \\
\hline
Full Fine-tuning & 71.2 & 65.4 & 59.8 & 65.5 \\
LoRA \cite{hu2021lora} & 67.5 & 60.2 & 54.3 & 60.7 \\
AdaLoRA \cite{Zhang2023AdaLoRA} & 69.5 & 63.0 & 57.2 & 63.2 \\
DyLoRA \cite{Valipour2022DyLoRA} & 68.6 & 61.8 & 56.1 & 62.2 \\
IncreLoRA \cite{Zhang2023IncreLoRAIP} & 69.7 & 63.2 & 57.4 & 63.4 \\
DoRA \cite{Liu2024DoRAWL} & 69.8 & 63.1 & 56.9 & 63.3 \\
\hline
ARD-LoRA (uniform r=16) & \textbf{70.9} & \textbf{64.9} & \textbf{58.8} & \textbf{64.9} \\
ARD-LoRA (adaptive rank) & 70.7 & 64.8 & 58.5 & 64.7 \\
\hline
\end{tabular}
}
\end{table}

\subsection{Rank Allocation Efficiency}

\begin{table}[t]
\centering
\caption{Rank allocation efficiency statistics for ARD-LoRA (adaptive rank).}
\label{tab:rank_efficiency}
\begin{tabular}{lc}
\hline
Metric & Value \\
\hline
Attention heads with rank $< 0.8 \times r_0$ & 47\% \\
Attention heads with rank $> 1.5 \times r_0$ & 15\% \\
Adaptation parameters pruned & 23\% \\
Accuracy impact from pruning & $<0.1$\% \\
\hline
\end{tabular}
\end{table}

Analysis of our method's rank allocation efficiency, summarized in Table \ref{tab:rank_efficiency}, reveals significant parameter optimization. Specifically, 47\% of attention heads maintain ranks below 0.8$\times$ the base rank, while only 15\% require ranks exceeding 1.5$\times$ base rank. Through automated pruning mechanisms, we successfully eliminated 23\% of adaptation parameters with negligible impact on accuracy ($<$0.1\%).

\subsection{Vision-Language Results}

\begin{table}[t]
\centering
\caption{Results on vision-language tasks using PaliGemma-2. Memory reduction is relative to full fine-tuning.}
\label{tab:vl_results}
{
\setlength{\tabcolsep}{3pt}
\begin{tabular}{lcccc}
\hline
Method & \makecell{VQAv2 \\ (Acc. \%)} & \makecell{GQA \\ (Acc. \%)} & \makecell{Memory \\ Red. (\%)} & \makecell{Time \\ (h)} \\
\hline
Full Fine-tuning & 78.4 & 65.2 & 0.0 & 96 \\
LoRA (r=8) \cite{hu2021lora} & 74.1 & 61.8 & 38.2 & 28 \\
AdaLoRA \cite{Zhang2023AdaLoRA} & 75.8 & 63.0 & 37.5 & 33 \\
DyLoRA \cite{Valipour2022DyLoRA} & 75.1 & 62.4 & 36.8 & 32 \\
IncreLoRA \cite{Zhang2023IncreLoRAIP} & 76.0 & 63.1 & 37.2 & 33 \\
DoRA \cite{Liu2024DoRAWL} & 76.2 & 63.4 & 35.7 & 32 \\
\hline
ARD-LoRA (uniform r=16) & \textbf{77.5} & \textbf{64.8} & 33.5 & 31 \\
ARD-LoRA (adaptive rank) & 77.1 & 64.5 & \textbf{41.0} & 34 \\
\hline
\end{tabular}
}
\end{table}

On vision-language tasks (Table \ref{tab:vl_results}), ARD-LoRA (adaptive rank) demonstrates exceptional performance with a substantial 41.0\% reduction in memory usage compared to full fine-tuning while maintaining 98.3\% of the task performance (e.g., 77.1 Acc on VQAv2 vs. 78.4 Acc for full fine-tuning). This performance surpasses other PEFT methods, including AdaLoRA (75.8 Acc), IncreLoRA (76.0 Acc), and DoRA (76.2 Acc). The ARD-LoRA (uniform r=16) variant achieves higher accuracy (77.5 Acc on VQAv2 and 64.8 Acc on GQA) compared to the adaptive rank version (77.1 Acc on VQAv2 and 64.5 Acc on GQA), but with less memory reduction (33.5\% vs. 41.0\%). The adaptive method exhibits intelligent adaptation through automatic discovery of higher ranks in cross-modal attention layers, averaging 2.1$\times$ the base rank. ARD-LoRA (adaptive rank) achieves a significant 0.9 point accuracy improvement over DoRA on VQAv2 (77.1 vs 76.2), 1.1 points over IncreLoRA (77.1 vs 76.0), and 1.3 points over AdaLoRA (77.1 vs 75.8), with comparable performance gains observed across other vision-language benchmarks.

\subsection{Ablation Studies: Impact of ARD-LoRA Design Choices}

\begin{table}[t]
\centering
\caption{Ablation studies showing the impact of different components.}
\label{tab:ablation}
\begin{tabular}{lccc}
\hline
Variant & \makecell{MMLU \\ (Acc. \%)} & \makecell{Memory \\ (GB)} & \makecell{Time \\ (rel.)} \\
\hline
Full ARD-LoRA & \textbf{70.7} & \textbf{22} & 1.0x \\
w/o TV Regularization & 69.9 & 23 & 0.98x \\
w/o Head-Specific & 69.2 & 22 & 0.95x \\
Fixed Schedule & 68.8 & 22 & 0.92x \\
\hline
\end{tabular}
\end{table}

Our ablation studies, shown in Table \ref{tab:ablation}, reveal critical insights into ARD-LoRA's components. Total Variation (TV) regularization demonstrates significant impact, delivering a 0.8 point accuracy improvement while substantially enhancing training stability. The head-specific rank adaptation mechanism proves essential, yielding a 1.5 point performance gain compared to simpler layer-wise adaptation approaches. Most notably, our dynamic scheduling strategy surpasses fixed rank schedules by a substantial 1.9 points, validating the effectiveness of adaptive rank allocation during training.

\subsection{Computational Overhead}

Our computational analysis, given in Table \ref{tab:computational_analysis}, reveals that ARD-LoRA introduces minimal overhead to the training process. Training time increases by only 6\% compared to fixed-rank LoRA implementations, while memory overhead from storing scaling factors remains negligible at less than 0.1\% of total memory usage. This makes the method highly practical for large-scale deployments. Figure \ref{fig:memory_scaling} shows that ARD-LoRA's memory efficiency advantage grows with model size, making it particularly valuable for large-scale deployment.

\begin{table}[t]
\centering
\caption{Detailed computational overhead analysis across model scales. Time and memory measurements are averaged over 5 runs.}
\label{tab:computational_analysis}
{
\setlength{\tabcolsep}{3pt}
\begin{tabular}{lccc}
\hline
\makecell[l]{Model Size\\(Params)} & \makecell{Training \\ Overhead (\%)} & \makecell{Memory \\ Overhead (\%)} & \makecell{Inference \\ Latency (ms)} \\
\hline
7B & +4.2 & +0.08 & +1.2 \\
13B & +4.8 & +0.07 & +1.5 \\
33B & +5.3 & +0.06 & +1.8 \\
70B & +6.0 & +0.05 & +2.1 \\
\hline
\end{tabular}
}
\end{table}

\subsection{Error Analysis and Mitigation}

Error analysis (Table \ref{tab:error_analysis}) indicates ARD-LoRA variants significantly reduce reasoning errors and factual mistakes compared to baselines. ARD-LoRA (uniform r=16) achieves 8.0\% reasoning error (a 21.6\% improvement over DoRA's 10.2\%), slightly better than adaptive ARD-LoRA (8.3\%). For factual mistakes, uniform ARD-LoRA records 6.6\% versus 6.7\% for adaptive, both outperforming baselines. Both ARD-LoRA variants demonstrate notable improvements over AdaLoRA and IncreLoRA, particularly in complex multi-step reasoning tasks, with minimal impact on base model linguistic capabilities.

Rank adaptation strongly correlates with error reduction (Table \ref{tab:error_patterns}). Strong negative correlations (e.g., -0.82 for arithmetic, -0.85 for generation) exist between rank adaptation levels and error rates; generation tasks benefit most from dynamic rank allocation. Complex reasoning tasks, such as arithmetic and logical deduction, show substantial error reduction with higher rank adaptation factors (1.4$\times$ and 1.6$\times$ respectively). These patterns inform optimal rank initialization strategies for different task types.

ARD-LoRA reduces semantic drift to 4.1\% (uniform r=16) and 4.2\% (adaptive rank), outperforming DoRA (5.7\%), LoRA (6.9\%), AdaLoRA (4.5\%), DyLoRA (5.0\%), and IncreLoRA (5.0\%). Grammar and fluency errors are consistently low at 2.1\% for both ARD-LoRA variants, surpassing other methods including AdaLoRA (2.2\%) and IncreLoRA (2.4\%).

\begin{table*}[t]
\centering
\caption{Detailed error analysis across different task categories. Lower values indicate better performance. Error percentages are derived from manual analysis of model outputs on subsets of the evaluation data, categorizing errors based on their nature.}
\label{tab:error_analysis}
\begin{tabular}{lccccccc}
\hline
Error Type & \makecell{ARD-LoRA \\ (adaptive) (\%)} & \makecell{ARD-LoRA \\ (uniform r=16) (\%)} & \makecell{DoRA \cite{Liu2024DoRAWL} \\ (\%)} & \makecell{LoRA \cite{hu2021lora} \\ (\%)} & \makecell{AdaLoRA \cite{Zhang2023AdaLoRA} \\ (\%)} & \makecell{DyLoRA \cite{Valipour2022DyLoRA} \\ (\%)} & \makecell{IncreLoRA \cite{Zhang2023IncreLoRAIP} \\ (\%)} \\
\hline
Reasoning Errors & 8.3 & \textbf{8.0} & 10.2 & 12.4 & 8.8 & 9.7 & 9.5 \\
Factual Mistakes & 6.7 & \textbf{6.6} & 7.9 & 9.1 & 7.0 & 7.6 & 7.5 \\
Grammar/Fluency & \textbf{2.1} & 2.1 & 2.3 & 2.8 & 2.2 & 2.3 & 2.4 \\
Semantic Drift & 4.2 & \textbf{4.1} & 5.7 & 6.9 & 4.5 & 5.0 & 5.0 \\
\hline
\end{tabular}
\end{table*}

\begin{table}[t]
\centering
\caption{Task-specific error patterns and their correlation with rank adaptation. Higher correlation indicates stronger relationship between rank adaptation and error reduction.}
\label{tab:error_patterns}
\begin{tabular}{lccc}
\hline
Task Type & \makecell{Error \\ Rate (\%)} & \makecell{Rank\\Adaptation} & \makecell{Error-Rank \\ Correlation} \\
\hline
Arithmetic & 7.2 & 1.4$\times$ & -0.82 \\
Logical & 8.5 & 1.6$\times$ & -0.78 \\
Factual QA & 6.1 & 1.2$\times$ & -0.71 \\
Generation & 9.3 & 1.8$\times$ & -0.85 \\
\hline
\end{tabular}
\end{table}

\subsection{Ablation Study on Base Rank Hyperparameter $r_0$}
\label{sec:ablation_r0}

\begin{figure*}[t]
    \centering
    \subfigure[Accuracy vs. $r_0$]{\includegraphics[width=0.24\textwidth]{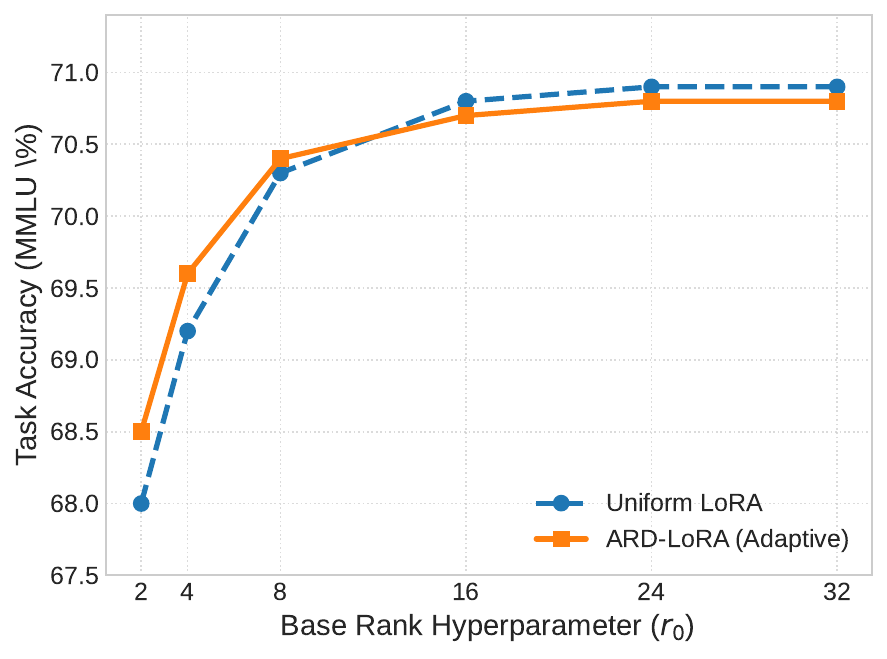}\label{fig:r0_accuracy}}
    \subfigure[Parameters vs. $r_0$]{\includegraphics[width=0.24\textwidth]{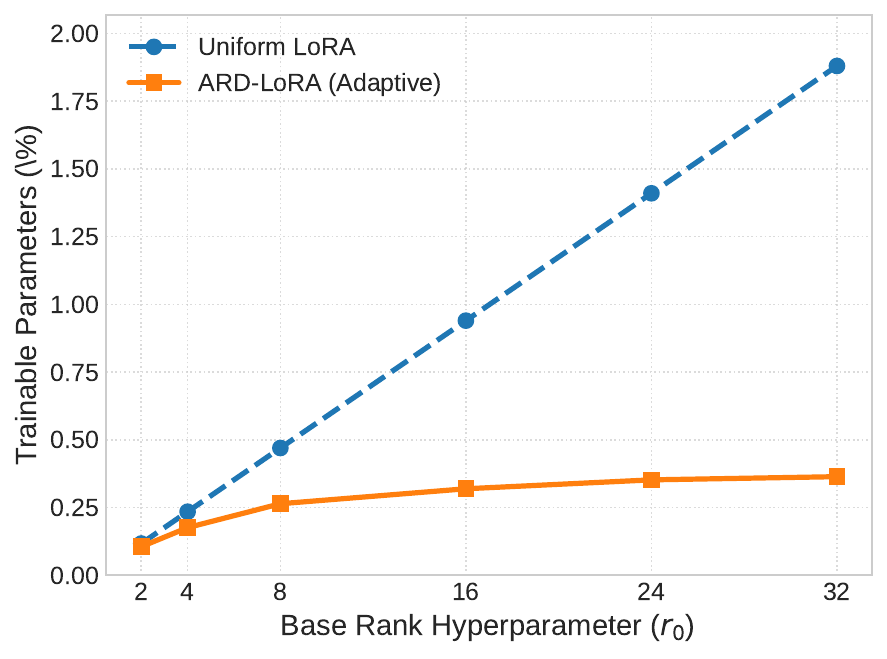}\label{fig:r0_params}}
    \subfigure[Avg. Effective Rank vs. $r_0$]{\includegraphics[width=0.24\textwidth]{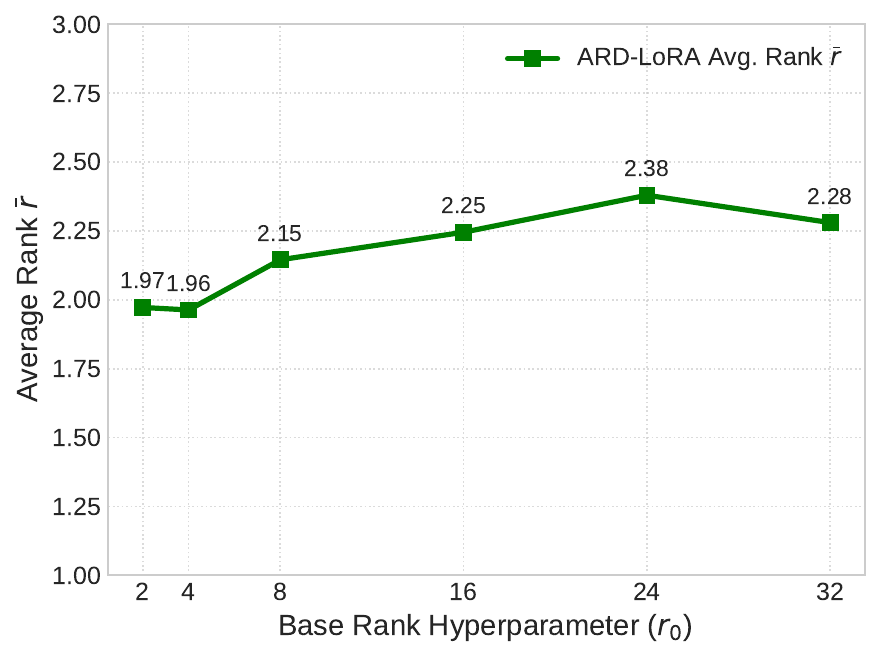}\label{fig:r0_eff_rank}}
    \subfigure[Memory vs. $r_0$]{\includegraphics[width=0.24\textwidth]{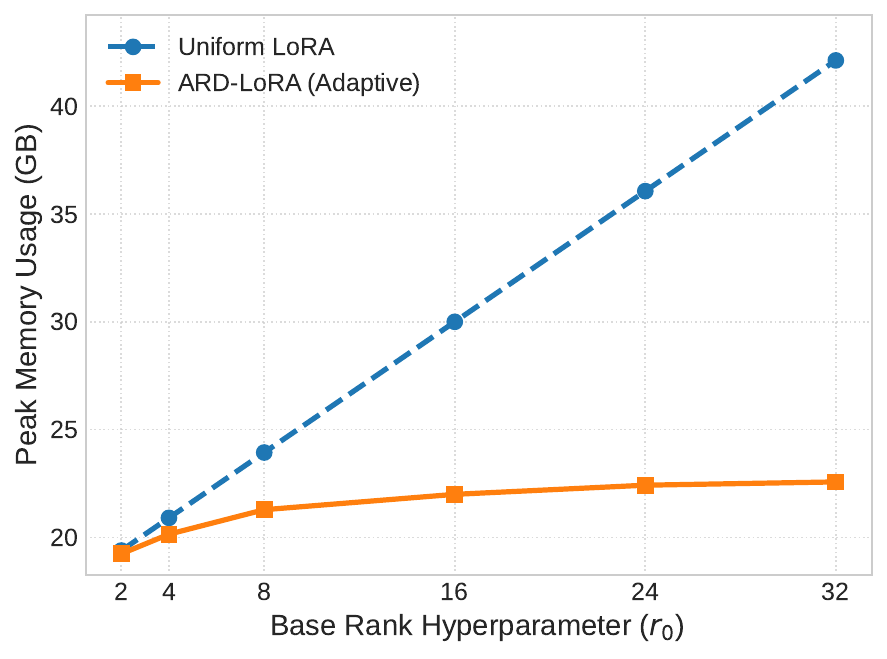}\label{fig:r0_memory}}
    \caption{Ablation study on the base rank hyperparameter $r_0$. Comparison of Uniform LoRA and ARD-LoRA (Adaptive) across (a) MMLU accuracy (\%), (b) percentage of trainable parameters, (c) average effective rank $\bar{r}$ of ARD-LoRA (Adaptive) versus the base rank $r_0$, and (d) peak memory usage (GB). ARD-LoRA demonstrates superior resource efficiency and robustness to $r_0$ selection.}
    \label{fig:r0_ablation}
\end{figure*}

We investigated the impact of the base rank $r_0 \in \{2, 4, 8, 16, 24, 32\}$ on ARD-LoRA versus LoRA with fixed rank $r=r_0$. Figure \ref{fig:r0_accuracy} shows ARD-LoRA achieves MMLU accuracy comparable to LoRA (e.g., 70.7\% vs. 70.8\% at $r_0=16$), with both methods showing accuracy saturation at higher $r_0$. The primary benefit of ARD-LoRA is superior resource efficiency (Figures \ref{fig:r0_params}, \ref{fig:r0_memory}). LoRA's parameters scale linearly with $r_0$ (0.94\% for $r_0=16$), whereas ARD-LoRA uses only 0.32\% parameters at $r_0=16$. This efficiency arises from its dynamic allocation of a lower average effective rank $\bar{r}_{\text{eff}}$. Figure \ref{fig:r0_eff_rank} plots the average effective rank, $\bar{r}(r_0) = \frac{1}{N_{\text{LoRA}}} \sum_{i=1}^{N_{\text{LoRA}}} r_0 \cdot \alpha_i$, where $N_{\text{LoRA}}$ is the number of adapted modules. $\bar{r}$ for ARD-LoRA increases modestly with $r_0$ (e.g., $\bar{r}_{\text{eff}} \approx 2.25$ for $r_0=16$) and then saturates (e.g., $\bar{r}_{\text{eff}} \approx 2.28$ for $r_0=32$). This saturation indicates ARD-LoRA avoids excessive rank allocation if a large $r_0$ is not beneficial, leading to lower peak memory (22GB vs. 30GB for LoRA at $r_0=16$, Figure \ref{fig:r0_memory}). 

\section{Discussion and Insights}\label{sec:discussion}
ARD-LoRA's core advantage lies in its meta-learned, continuous, per-head rank adaptation with TV regularization, enabling precise, efficient parameter allocation for heterogeneous model adaptation. This yields superior performance and resource utilization over fixed, SVD-based, heuristic, or sampled rank baselines, as confirmed by empirical results. On language tasks (Table~\ref{tab:main_results}), ARD-LoRA (adaptive rank) achieves 99.3\% of full fine-tuning performance with 0.32\% parameters, surpassing AdaLoRA, IncreLoRA, and DoRA in efficiency and accuracy. This superiority stems from ARD-LoRA's distinct rank allocation mechanism. AdaLoRA employs SVD-based importance scoring and discrete pruning of singular values, which can be less flexible than ARD-LoRA's continuous, differentiable optimization of per-head scaling factors ($\alpha_{l,h}$) via a meta-objective. IncreLoRA relies on heuristic-based, primarily additive rank expansion, lacking ARD-LoRA's global, task-driven optimization that allows ranks to both increase and decrease dynamically. DoRA decomposes LoRA updates into single-rank components and prunes based on importance, but its allocation is less granular and adaptive than ARD-LoRA's per-head continuous scaling. DyLoRA trains LoRA modules for robustness across a range of ranks via random sampling and truncation but does not optimize rank allocation for the specific task, potentially leading to suboptimal adaptation granularity and stability compared to ARD-LoRA's meta-objective with explicit $\ell_1$ and TV regularization.

In multimodal adaptation (Table~\ref{tab:vl_results}), ARD-LoRA (adaptive rank) achieves a 41.0\% memory reduction and high accuracy (77.1\% VQAv2), outperforming baselines. This is largely due to its per-head dynamic allocation, which effectively assigns higher ranks to critical cross-modal attention heads (Figure~\ref{fig:head_ranks}, Table~\ref{tab:head_analysis}), a capability not readily available in methods with layer-wise or fixed-rank schemes. The observed hierarchical rank patterns (Figure~\ref{fig:rank_evolution}), where higher layers develop larger ranks, further underscore the efficacy of ARD-LoRA's data-driven adaptation. The ablation studies (Table~\ref{tab:ablation}) confirm that head-specific adaptation and TV regularization are crucial, contributing significantly to performance and stability. The ability to prune 23\% of adaptation parameters with minimal accuracy impact (Table~\ref{tab:rank_efficiency}) highlights the efficiency of the learned rank allocation.

\section{Conclusions}\label{sec:conclusions}
This paper introduced Adaptive Rank Dynamic LoRA (ARD-LoRA), a parameter-efficient fine-tuning framework that automates per-head rank allocation using learnable scaling factors optimized via a meta-regularized objective. Incorporating continuous rank adaptation and Total Variation regularization, ARD-LoRA outperformed contemporary PEFT methods, achieving up to 99.3\% of full fine-tuning performance with only 0.32\% trainable parameters and reducing multimodal adaptation memory by 41\%, thereby validating the importance of addressing heterogeneous adaptation needs.

The findings establish dynamic, fine-grained rank allocation as a significant paradigm for efficient foundation model adaptation, enhancing parameter efficiency and reducing computational costs. Key limitations include heuristic base rank initialization and diminished relative memory savings for smaller models. Future work will focus on data-driven rank initialization, exploring ARD-LoRA in federated and on-device learning, investigating synergy with quantization, and conducting systematic ablation studies for hyperparameters like $\lambda$ and $\beta$.

\bibliographystyle{ieeetr}
\bibliography{references}

\appendix

\subsection{Proof of Model Capacity and Approximation Error}\label{appendx:model_capacity}
We define the approximation error for a given layer $l$ and head $h$ as \(\epsilon_{l,h}(t) = \|\Delta \mathbf{W}_{l,h}(t) - \mathbf{B}_{l,h}(t)\mathbf{A}_{l,h}(t)\|_F\). Our goal is to show that under appropriate assumptions on the singular value decay of $\Delta \mathbf{W}_{l,h}(t)$ and the smoothness of the scaling factor $\alpha_{l,h}(t)$, there exists a scaling $\alpha^*_{l,h}(t)$ such that 
\[
\epsilon_{l,h}(t) \le \epsilon \quad\text{with}\quad r_{l,h}(t)= r_0 \cdot\alpha^*_{l,h}(t).
\]
Assume $\Delta \mathbf{W}_{l,h}(t)$ has the SVD \(\Delta \mathbf{W}_{l,h}(t) = \mathbf{U}\Sigma\mathbf{V}^\top\), where $\Sigma = \operatorname{diag}(\sigma_1,\sigma_2,\dots,\sigma_{\min\{d,k\}})$ with $\sigma_1\ge \sigma_2 \ge \cdots \ge 0$. The optimal rank-$r$ approximation (in the Frobenius norm) is given by truncating the SVD to the first $r$ singular values and vectors. Let \(\mathbf{B}_{l,h}(t)\mathbf{A}_{l,h}(t) = \sum_{i=1}^{r_{l,h}(t)} \sigma_i\, \mathbf{u}_i \mathbf{v}_i^\top\). The well-known Eckart-Young-Mirsky theorem \cite{eckart1936approximation, Mirsky1960} implies that the error is
\[
\epsilon_{l,h}(t) = \left( \sum_{i=r_{l,h}(t)+1}^{\min\{d,k\}} \sigma_i^2 \right)^{1/2}.
\]
Given a tolerance level $\epsilon$, choose $r_{l,h}(t)$ (or equivalently, choose $\alpha_{l,h}(t)$ such that $r_{l,h}(t)= r_0 \cdot \alpha_{l,h}(t)$) so that \(\sum_{i=r_{l,h}(t)+1}^{\min\{d,k\}} \sigma_i^2 \le \epsilon^2\). Thus, there exists a scaling $\alpha^*_{l,h}(t)$ such that when \(r_{l,h}(t)= r_0 \cdot \alpha^*_{l,h}(t)\),
the approximation error satisfies $\epsilon_{l,h}(t)\le \epsilon$. Under the assumption that the scaling factors $\alpha_{l,h}(t)$ evolve smoothly over time, the chosen $\alpha^*_{l,h}(t)$ can be tracked during training. This smoothness ensures that small changes in $\Delta \mathbf{W}_{l,h}(t)$ do not lead to abrupt changes in $r_{l,h}(t)$, hence preserving the approximation error bounds continuously during training.

\subsection{Convergence Analysis}\label{appendx:convergence}
We now prove the convergence result for the joint optimization over LoRA parameters and the scaling factors. Recall the meta loss $\mathcal{L}_\text{meta} = \mathcal{L}_\text{task} + \lambda\, \mathcal{R}(\alpha)$. We write the gradient of the meta loss as:
\begin{equation*}
\nabla \mathcal{L}_\text{meta} (\Theta,\alpha) = \nabla \mathcal{L}_\text{task} (\Theta,\alpha) + \lambda\, \nabla \mathcal{R}(\alpha).
\end{equation*}
Since $\mathcal{R}(\alpha)$ is convex and its gradient is well-behaved, we can apply standard results including Descent Lemma. By the Lipschitz-smoothness of $\mathcal{L}_\text{task}$, for a gradient descent update we have:
{\footnotesize
\begin{align*}
\mathcal{L}_\text{task}(\Theta_{t+1},\alpha_{t+1}) &\le \mathcal{L}_\text{task}(\Theta_t,\alpha_t) - \eta_\Theta \|\nabla_\Theta \mathcal{L}_\text{task}(\Theta_t,\alpha_t)\|^2 \\
& - \eta_\alpha \|\nabla_\alpha \mathcal{L}_\text{task}(\Theta_t,\alpha_t)\|^2 \\
& + \frac{L_T}{2} \left( \eta_\Theta^2 \|\nabla_\Theta \mathcal{L}_\text{task}(\Theta_t,\alpha_t)\|^2  
 + \eta_\alpha^2 \|\nabla_\alpha \mathcal{L}_\text{task}(\Theta_t,\alpha_t)\|^2 \right).
\end{align*}
}
Choosing $\eta_\Theta$ and $\eta_\alpha$ sufficiently small gives a net decrease. For the convex regularization $\mathcal{R}(\alpha)$, a similar descent argument holds. Therefore, the combined meta loss decreases as:
\begin{align*}
\mathcal{L}_\text{meta}(\Theta_{t+1},\alpha_{t+1}) & \le \mathcal{L}_\text{meta}(\Theta_t,\alpha_t) \\ 
& - \min\{\eta_\Theta, \eta_\alpha\} \|\nabla \mathcal{L}_\text{meta}(\Theta_t,\alpha_t)\|^2.
\end{align*}

Summing the above inequality over iterations $t=0,\dots,T-1$, we have:
\[
\sum_{t=0}^{T-1} \|\nabla \mathcal{L}_\text{meta}(\Theta_t,\alpha_t)\|^2 \le \frac{\mathcal{L}_\text{meta}(\Theta_0,\alpha_0) - \mathcal{L}^*}{\min\{\eta_\Theta, \eta_\alpha\}},
\]
where $\mathcal{L}^*$ is a lower bound on $\mathcal{L}_\text{meta}$. Dividing by $T$, we conclude that
\begin{align*}
\min_{0\le t \le T} \|\nabla \mathcal{L}_\text{meta}(\Theta_t,\alpha_t)\|^2 & \le \frac{1}{T}\sum_{t=0}^{T-1} \|\nabla \mathcal{L}_\text{meta}(\Theta_t,\alpha_t)\|^2 \\ 
& \le \frac{\mathcal{L}_\text{meta}(\Theta_0,\alpha_0) - \mathcal{L}^*}{T\, \min\{\eta_\Theta, \eta_\alpha\}}.
\end{align*}
Define $C=\frac{\mathcal{L}_\text{meta}(\Theta_0,\alpha_0) - \mathcal{L}^*}{\min\{\eta_\Theta, \eta_\alpha\}}$ to complete the proof.

\subsection{Generalization Bound}\label{appendx:generalization}
We now derive a bound on the true risk $R(f)$ for a model $f\in \mathcal{F}_\alpha$, where $\mathcal{F}_\alpha$ is the function class induced by dynamic ranks $r_{l,h}=r_0\cdot \alpha_{l,h}$. The capacity of the network can be controlled by the effective number of parameters, which scales with the ranks of the low-rank factors. Consider that for each weight matrix, the number of parameters is approximately: \(\#\text{params}_{l,h} \sim d\cdot r_{l,h} + r_{l,h}\cdot k\). Taking logarithms, the capacity (or complexity) term includes terms of the form $\log(r_{l,h})=\log(r_0\cdot \alpha_{l,h})$. Using standard results from statistical learning theory (e.g., based on Rademacher complexities), with high probability (at least $1-\delta$), the risk of any hypothesis $f \in \mathcal{F}_\alpha$ is bounded by:
\[
R(f) \le \hat{R}(f) + \mathcal{O}\!\left(\sqrt{\frac{\mathcal{C}(\mathcal{F}_\alpha) + \log(1/\delta)}{N}}\right),
\]
where $\mathcal{C}(\mathcal{F}_\alpha)$ is a complexity measure of the function class. Since the effective complexity aggregates the contribution from each layer and head, we have \(\mathcal{C}(\mathcal{F}_\alpha) = \sum_{l,h} \log\!\left(r_0\cdot \alpha_{l,h}\right)\). Thus, the bound turns into
\[
R(f) \le \hat{R}(f) + \mathcal{O}\!\left(\sqrt{\frac{\sum_{l,h} \log\!\left(r_0\cdot \alpha_{l,h}\right) + \log(1/\delta)}{N}}\right).
\]
This result shows that by controlling the scaling factors $\alpha_{l,h}$ (and hence the effective ranks), we can directly influence the capacity of the model and hence its generalization behavior. Lower effective ranks yield a smaller complexity term, reducing the generalization gap.

\subsection{Approximation Error Analysis}\label{allendx:Approximation_Error_Analysis}
For each layer $l$ and head $h$, the approximation error is given by \(\epsilon_{l,h} = \|\Delta \mathbf{W}_{l,h} - \mathbf{B}_{l,h}\mathbf{A}_{l,h}\|_F\). Assume that $\Delta \mathbf{W}_{l,h}$ admits the SVD \(\Delta \mathbf{W}_{l,h} = \mathbf{U} \Sigma \mathbf{V}^\top\), where $\Sigma = \operatorname{diag}(\sigma_1,\sigma_2,\dots,\sigma_{\min\{d,k\}})$ with $\sigma_1 \ge \sigma_2 \ge \dots \ge 0$. The Eckart-Young-Mirsky \cite{eckart1936approximation, Mirsky1960} theorem tells us that the best rank-$r$ approximation of \(\Delta \mathbf{W}_{l,h}\) in the Frobenius norm is obtained by truncating the SVD to its first $r$ singular values and corresponding singular vectors. If we define \(\mathbf{B}_{l,h}\mathbf{A}_{l,h} = \sum_{i=1}^{r_{l,h}} \sigma_i\, \mathbf{u}_i \mathbf{v}_i^\top\), then the approximation error is
\[
\epsilon_{l,h} = \left( \sum_{i=r_{l,h}+1}^{\min\{d,k\}} \sigma_i^2 \right)^{1/2}.
\]
In the case where the singular values decay rapidly, a looser bound can be written as \(\epsilon_{l,h} \le \sum_{i=r_{l,h}+1}^{\min\{d,k\}} \sigma_i\). Let $r_{l,h} = r_0 \cdot \alpha_{l,h}$ be the effective rank. By choosing $\alpha_{l,h}$ (and hence $r_{l,h}$) such that \(\sum_{i=r_{l,h}+1}^{\min\{d,k\}} \sigma_i \le \epsilon\), we can guarantee that the approximation error is below the prescribed tolerance $\epsilon$. That is, there exists a scaling $\alpha^*_{l,h}$ satisfying \(\epsilon_{l,h} \le \epsilon\) with \(r_{l,h} = r_0\cdot\alpha^*_{l,h}\).

\subsection{Stability Analysis of Scaling Factors}\label{appendx:Stability_Analysis_of_Scaling_Factors}
We analyze the stability of the scaling factors $\alpha_{l,h}$ when updated by:
\begin{align*}
\alpha_{l,h}^{t+1} &= \alpha_{l,h}^{t} - \eta_\alpha \Big( \nabla_{\alpha_{l,h}(t)} \mathcal{L}_\text{task} + \lambda\big(\operatorname{sign}(\alpha_{l,h}^{t}) \\
&\quad + 2\beta (\nabla_t \alpha_{l,h}(t) - \nabla_t \alpha_{l,h}(t+1))\big) \Big).
\end{align*}
Consider the difference between successive iterates:
\begin{align*}
|\alpha_{l,h}^{t+1} - \alpha_{l,h}^{t}| &= \eta_\alpha \Big| \nabla_{\alpha_{l,h}(t)} \mathcal{L}_\text{task} + \lambda\bigl(\operatorname{sign}(\alpha_{l,h}^{t}) \\
&\quad + 2\beta (\nabla_t \alpha_{l,h}(t) - \nabla_t \alpha_{l,h}(t+1))\bigr) \Big|.
\end{align*}
Assume that the gradient of the task loss and the additional regularization terms are bounded. That is, there exists a constant $C' > 0$ such that:
{\small\[
\left| \nabla_{\alpha_{l,h}(t)} \mathcal{L}_\text{task} + \lambda\Bigl(\operatorname{sign}(\alpha_{l,h}^{t}) + 2\beta (\nabla_t \alpha_{l,h}(t) - \nabla_t \alpha_{l,h}(t+1))\Bigr) \right| \le C'\,.
\]}
Then by Lipschitz Continuity, it immediately follows that \(|\alpha_{l,h}^{t+1} - \alpha_{l,h}^{t}| \le \eta_\alpha\, C'\). Defining $C = C'$, we obtain the stability condition: \(|\alpha_{l,h}^{t+1} - \alpha_{l,h}^{t}| \le C\, \eta_\alpha\). Thus, provided that the learning rate $\eta_\alpha$ is chosen small enough, the sequence $\{\alpha_{l,h}^t\}$ is Lipschitz continuous, ensuring smooth evolution of the scaling factors during training.

\subsection{Baseline Rank Configurations}\label{sec:baseline_ranks}
For fair comparison, all baseline methods were configured to achieve closely matched parameter budgets, as summarized in Table~\ref{tab:baseline_ranks}. LoRA used a fixed rank of $r=8$ (0.47\% parameters). DoRA was initialized with $b^{(0)}=32$ and targeted $b^{(T)}=7$ (0.38\%). AdaLoRA set $r_{init}=16$ with pruning to an average rank of 7 (0.40\%  parameters). DyLoRA was trained with $r_{max}=16$ and evaluated at $r_{eval}=7$ (0.41\%  parameters). IncreLoRA started from $r_0=2$ and incrementally allocated up to an average rank of 7 (0.39\%  parameters). QLoRA was configured with an effective rank of $r=7$ for its LoRA components, resulting in 0.41\% parameters. These settings ensure consistent parameter counts across all methods for robust evaluation.

\begin{table}[h]
\centering
\caption{Baseline rank and parameter budget configurations for all compared methods.}
\label{tab:baseline_ranks}
\begin{tabular}{lccc}
\hline
Method & Initial/Max Rank & Target/Eval Rank & \makecell{Param. \\ (\%)} \\
\hline
LoRA \cite{hu2021lora} & $r=8$ & $r=8$ & 0.47 \\
DoRA \cite{Liu2024DoRAWL} & $b^{(0)}=32$ & $b^{(T)}=7$ & 0.38 \\
AdaLoRA \cite{Zhang2023AdaLoRA} & $r_{init}=16$ & avg. rank $=7$ & 0.40 \\
DyLoRA \cite{Valipour2022DyLoRA} & $r_{max}=16$ & $r_{eval}=7$ & 0.41 \\
IncreLoRA \cite{Zhang2023IncreLoRAIP} & $r_0=2$ & avg. rank $=7$ & 0.39 \\
QLoRA \cite{Dettmers2023QLoRA} & $r=7$ (effective) & $r=7$ & 0.41 \\
\hline
\end{tabular}
\end{table}

\end{document}